\documentclass{article}
\usepackage{mlsys2025}
\renewcommand{\icmlruler}[1]{}
\usepackage{etoolbox}
\makeatletter
\patchcmd{\mlsysauthor}{\ifdefined\isaccepted}{\iftrue}{}{%
  \PackageError{authordetails}{Could not show author names}{Check the MLSys style.}}
\patchcmd{\mlsysaffiliation}{\ifdefined\isaccepted}{\iftrue}{}{%
  \PackageError{authordetails}{Could not show affiliations}{Check the MLSys style.}}
\patchcmd{\mlsyscorrespondingauthor}{\ifdefined\isaccepted}{\iftrue}{}{%
  \PackageError{authordetails}{Could not show the corresponding author}{Check the MLSys style.}}
\makeatother
\usepackage[T1]{fontenc}
\usepackage[utf8]{inputenc}
\usepackage{microtype}
\usepackage{hyperref,url}
\usepackage{xurl}
\usepackage{amsmath,amssymb,mathtools}
\usepackage{booktabs,tabularx,array,multirow}
\usepackage{graphicx,xcolor}
\newcolumntype{Y}{>{\raggedright\arraybackslash}X}

\usepackage{placeins}
\hypersetup{colorlinks=true,linkcolor=[rgb]{0.08,0.27,0.55},citecolor=[rgb]{0.08,0.27,0.55},urlcolor=[rgb]{0.08,0.27,0.55}}
\renewcommand{\algorithmiccomment}[1]{\hfill{\color{gray}$\triangleright$~#1}}

\mlsystitlerunning{Guarded Adam-State Compression with Shared-Gradient Bounds}

\hypersetup{pdfsubject={MLSys 2027 submission draft},pdfauthor={}}

\begin{document}

\twocolumn[

\mlsystitle{Guarded Compression of Adam Optimizer State with Shared-Gradient Bounds}

\begin{mlsysauthorlist}

\mlsysauthor{Annan Zhang}{pku}
\mlsysauthor{Shengjie Song}{pku}
\mlsysauthor{Wenshuai Yao}{pku}
\mlsysauthor{Lihao Zhang}{pku}
\mlsysauthor{Tianmiao}{buaa}

\end{mlsysauthorlist}

\mlsysaffiliation{pku}{Peking University}
\mlsysaffiliation{buaa}{Beihang University}

\mlsyscorrespondingauthor{Annan Zhang}{\mbox{annan@pku.edu.cn}}

\vskip 0.3in

\begin{abstract}

Adam keeps two FP32 moments per parameter, twice the bytes of the FP32 weights, yet compressing them is risky: the update divides by the square root of the second moment, so small reconstruction errors can change the restored computation substantially. We study \emph{guarded} moment compression, in which a coordinate's decoded moments are accepted only if they reproduce the native FP32 Adam update to within a tolerance $\tau$, in units of the learning rate, for every gradient in a clipped domain; all other coordinates are stored exactly. The natural certificate, which ranges each update separately over gradient intervals, fails: on one million trained OPT-125M coordinates it escapes 98.9\% of coordinates and the payload grows to 126.6\% of raw FP32. Our key observation is that both updates consume the same gradient and parameter. Bounding their difference directly, with FP32 rounding-error allowances and a fixed mixed linear--logarithmic gradient partition, cuts escapes to 0.53\% and the payload to 28.3\% at $\tau=0.01$. Across 120 complete-checkpoint restores of OPT-125M and ResNet-18, with four trained seeds and three training fractions per model, guarded 8-bit log codes shrink moment payloads by 69.5\% and 71.9\% relative to FP32, 40.8\% and 45.4\% relative to equally guarded BF16, and 29.3\% and 37.6\% relative to equally guarded blockwise 8-bit, and every restored first update stays within tolerance. Encoding and restoration together take 2.3--2.6$\times$ as long as guarded BF16 but less time than guarded blockwise 8-bit. Searching both moment codes jointly saves under 0.001\% more bytes at 6\% more codec time, so per-moment rounding suffices. The guarantee covers the first restored update under a pinned Torch operation model; we measure subsequent training empirically.

\end{abstract}

]

\printAffiliationsAndNotice{}

\section{Introduction}
\label{sec:intro}
\begin{figure*}[t]
\centering
\includegraphics[height=4.2cm]{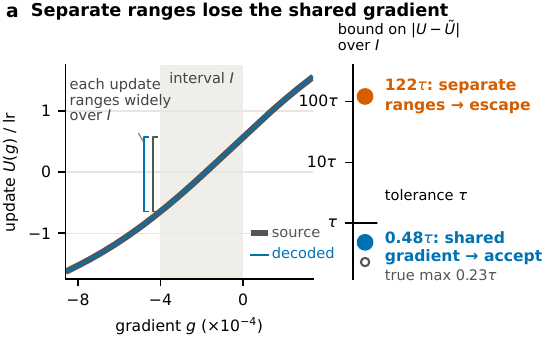}\hfill
\raisebox{0.05cm}{\includegraphics[height=4.2cm]{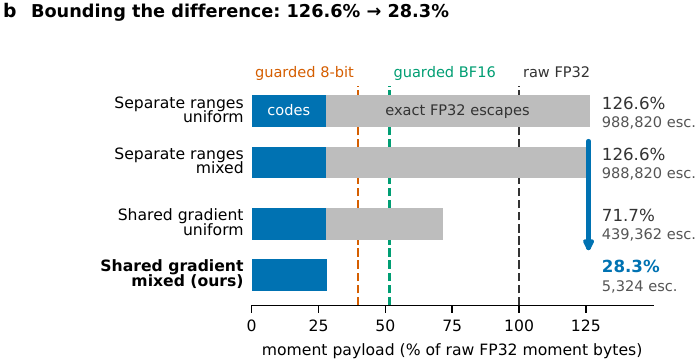}}
\caption{Bound the difference, not the outputs: the shared-gradient bound with a mixed partition shrinks the guarded payload from 126.6\% to 28.3\% of raw FP32, below equally guarded BF16 and blockwise 8-bit. \textbf{(a)}~Illustration with representative moments ($m=2\times10^{-5}$, $v=10^{-9}$, bias corrections $\approx1$) and decoded moments perturbed by $+0.2\%$ and $-0.4\%$. Over the gradient interval $I$, both updates sweep a range far wider than $\tau$, so bounding them separately gives $122\tau$; the shared-gradient bound of Equation~\eqref{eq:common} gives $0.48\tau$, against a true maximum of $0.23\tau$ (rounding allowances omitted). \textbf{(b)}~Measured on one million trained OPT-125M coordinates at $\tau=0.01$ with independent code selection. Every configuration spends the same 27.7\% of raw FP32 bytes on codes and metadata; the bound decides how many coordinates escape, at 8 bytes each. Vertical lines mark equally guarded BF16 and blockwise 8-bit payloads on the same coordinates.}
\label{fig:teaser}
\end{figure*}
Checkpoints are how long training runs survive failures, preemption and migration, and systems increasingly write them often~\citep{checkfreq,gemini}. For models trained with Adam~\citep{kingma2015adam}, the optimizer state dominates each checkpoint: two FP32 moments per parameter, twice the bytes of the FP32 weights. Every byte removed from the moments is removed from every checkpoint that is written, moved and stored.

Lossy formats are the obvious lever. BF16 moments halve the bytes, and blockwise 8-bit quantization~\citep{eightbit} quarters them. A restored optimizer, however, does not consume moments; it consumes the update they produce, and Adam divides the first moment by the square root of the second. Where the second moment is small, a small relative error changes the update substantially. On one million coordinates of a trained OPT-125M checkpoint, plain blockwise 8-bit moments change the next update by more than 1\% of a learning-rate step on 2,469 coordinates for a fixed set of in-domain gradients, and by up to 12.9\% (\S\ref{sec:baselines}). A checkpoint codec should therefore constrain the \emph{update} produced by the decoded moments, not the moments themselves.

We call this \emph{guarded} compression. Decoded moments for a coordinate are accepted only if, for every gradient $g$ in the clipping domain $|g|\le C$, the native FP32 Adam update they produce differs from the original by at most $\tau$ learning rates. Coordinates that fail are \emph{escaped}, that is, stored as exact FP32 moments, so the contract holds for every coordinate, and every code, escape and metadata byte is counted.

The difficulty is certifying a continuum of gradients. The textbook tool is interval arithmetic~\citep{moore2009}: partition the domain, bound the range of each update over each interval, and bound the difference of the updates by the distance between the two ranges. This discards the fact that both updates consume the same gradient, the classic \emph{dependency problem}. Near $g=0$ the update can sweep across most of its range within one interval, so the two ranges are wide even when the updates are nearly identical (Figure~\ref{fig:teaser}a). On the same million coordinates, this separate-range bound escapes 98.9\% of coordinates, and the ``compressed'' moments occupy 126.6\% of raw FP32, more than storing them uncompressed. Refining the gradient partition does not help (Figure~\ref{fig:teaser}b).

Our key observation is that the source and restored updates share their inputs, so the bound belongs on their \emph{difference}. Subtracting the two Adam quotients algebraically cancels the gradient from the numerator difference and the parameter from the update difference, leaving terms that scale with the moment reconstruction errors (\S\ref{sec:shared}). We add explicit rounding-error allowances for each native FP32 operation path and evaluate the bound over a fixed partition that mixes linear and base-2 logarithmic gradient magnitudes. Escapes fall from 988,820 to 5,324, and the payload falls to 28.3\% of raw FP32: 45\% smaller than BF16 and 29\% smaller than blockwise 8-bit moments under the same guard.

This paper makes three contributions:
\begin{itemize}\setlength{\itemsep}{1pt}\setlength{\parskip}{0pt}\setlength{\topsep}{2pt}
\item A \textbf{shared-gradient enclosure} for the difference between two native Adam updates, combined with FP32 rounding-error allowances under a pinned Torch operation contract (\S\ref{sec:design}).
\item A \textbf{guarded moment codec} with per-coordinate acceptance on decoded values, exact escapes, a fully accounted serialized format and standalone readback verification, implemented with directed-rounding CUDA kernels and audited on an H100 (\S\ref{sec:design}, \S\ref{sec:mechanism}).
\item An \textbf{evaluation that separates mechanism, baselines and cost} (\S\ref{sec:eval}): a controlled ablation; equally guarded BF16 and official bitsandbytes baselines; 120 complete-checkpoint restores of OPT-125M and ResNet-18 with 100-update continuation; and paired multi-process codec timing.
\end{itemize}

Across both models, guarded log codes reduce moment payloads by 67--72\% relative to FP32, and every first restored update passes its audit. Log codes also need 17--28\% less combined codec time than guarded blockwise 8-bit, and their combined codec time is 2.3--2.6$\times$ that of guarded BF16. The design is simple. Once the acceptance test is tight, rounding each moment independently captures essentially all of the savings; searching code pairs jointly adds less than 0.001\% while costing about 6\% more codec time.

The guarantee is precise: under the pinned operation model, every restored first update provably matches the native FP32 update to within $\tau$. Later updates consume gradients that differ between branches, and we measure them directly over 100-update continuations (\S\ref{sec:fidelity}, \S\ref{sec:limits}).

\section{Background and Problem}
\label{sec:problem}
For moments $(m,v)$ saved after $t$ completed updates and a next gradient $g$, Adam computes
\begin{align}
 m^+&=\beta_1m+(1-\beta_1)g,\\
 v^+&=\beta_2v+(1-\beta_2)g^2,\\
 U_{t+1}(m,v;g)&=
 \frac{m^+/(1-\beta_1^{t+1})}
 {\sqrt{v^+/(1-\beta_2^{t+1})}+\epsilon},
\end{align}
and the ideal parameter change is $-\eta U_{t+1}$. Because $v$ appears under a square root in the denominator, a small absolute error in a small $v$ can change $U_{t+1}$ by a large amount.

\subsection{The restored-update contract}
Let $\Phi(w,m,v;g)$ denote the native update program: the FP32 parameter that a pinned optimizer implementation produces from parameter $w$, moments $(m,v)$ and gradient $g$. Decoded moments $(\widetilde m,\widetilde v)$ satisfy the contract if
\begin{equation}
 \sup_{|g|\le C}
 \frac{|\Phi(w,m,v;g)-\Phi(w,\widetilde m,\widetilde v;g)|}{\eta}
 \le \tau .
 \label{eq:native-contract}
\end{equation}
Both branches share $w$, the realized gradient and all optimizer metadata, including the update counter $t$ supplied by the checkpoint. We use $C=0.2$, consistent with the gradient clipping of our training pipelines; a gradient outside the declared domain is outside the contract.

\paragraph{Operation contract.} Our native target is Torch~2.8~\citep{pytorch} scalar FP32 Adam with \texttt{foreach}, \texttt{fused}, \texttt{capturable}, \texttt{differentiable}, \texttt{amsgrad} and \texttt{maximize} disabled, zero weight decay (AdamW's decoupled decay~\citep{adamw} is out of scope), and real FP32 parameters, gradients and moments. The enclosure fixes $\beta_1=0.9$, $\beta_2=0.999$ and $\epsilon=10^{-8}$. It requires finite $|w|\le10^6$, $|m|,|\widetilde m|\le1$, $0\le v,\widetilde v\le1$, $C\le1$, a learning rate in $[10^{-6},0.01]$ and a supported integer update counter. Any other input stops certification.

\subsection{Why separate ranges fail}
\label{sec:separate}
A certificate for Equation~\eqref{eq:native-contract} must cover a continuum of gradients. Interval methods partition $[-C,C]$ into intervals $I$ and bound each output over each interval. Ranging the two outputs separately yields $|\Phi-\widetilde\Phi|\le\max(\Phi_{\rm hi}-\widetilde\Phi_{\rm lo},\,\widetilde\Phi_{\rm hi}-\Phi_{\rm lo})$, which can never be smaller than the variation of either update across $I$. That variation is largest exactly where precision matters: when $v$ is small, the update swings from strongly negative to strongly positive over a narrow band of gradients near zero (Figure~\ref{fig:teaser}a). The bound then measures how strongly the update depends on $g$, not how much the two moment pairs differ.

\section{Design}
\label{sec:design}

\begin{figure*}[t]
\centering
\includegraphics[width=\textwidth]{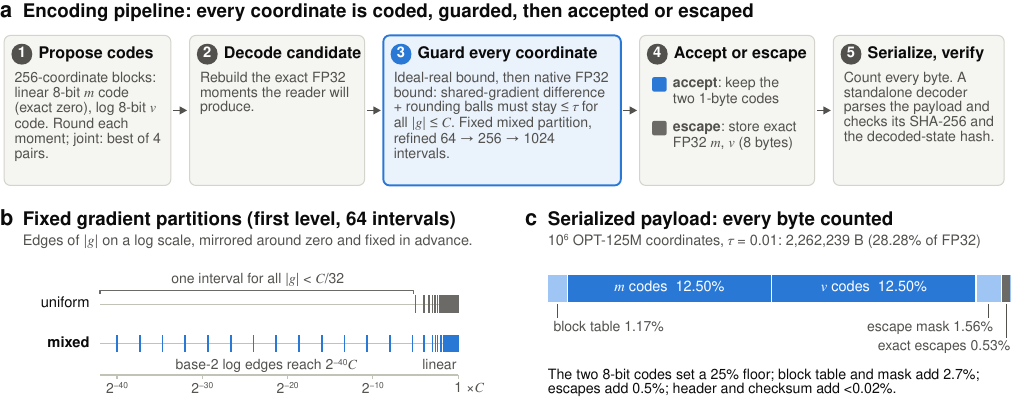}
\caption{Guarded codec design. \textbf{(a)}~The encoder proposes 8-bit codes per 256-coordinate block, decodes them exactly, and guards every coordinate with an ideal-real bound and the native shared-gradient bound; coordinates that fail are escaped to exact FP32. \textbf{(b)}~First-level edges of the two fixed partitions on a log-magnitude axis. A uniform partition covers all $|g|<C/32$ with one interval on each side of zero; the mixed partition adds base-2 logarithmic edges down to $2^{-40}C$. \textbf{(c)}~Byte layout of the serialized payload for the one-million-coordinate subset. The header and SHA-256 checksum (879 bytes in total) are too small to show.}
\label{fig:design}
\end{figure*}

\subsection{Overview}
Figure~\ref{fig:design}a summarizes the encoder. For each block of 256 coordinates it proposes an 8-bit code for each moment and decodes the candidate exactly as a reader would. It then guards every coordinate twice: first with an ideal-real bound computed with directed rounding, then with the native bound of \S\ref{sec:shared}--\ref{sec:balls}. Accepted coordinates keep their codes; the rest are escaped. The serialized payload (Figure~\ref{fig:design}c) holds a header, a per-block table, the two code arrays, an escape bitmask, the escaped FP32 pairs and a SHA-256 checksum. Before restoration, a standalone decoder parses the complete payload and must reproduce the hash of the decoded state.
Algorithm~\ref{alg:encode} lists the steps.

\begin{algorithm}[t]
\caption{Guarded encoding of one moment pair $(m,v)$.}
\label{alg:encode}
\begin{algorithmic}[1]
\REQUIRE FP32 $w,m,v\in\mathbb{R}^n$; counter $t$; learning rate $\eta$; domain $C$; tolerance $\tau$
\ENSURE serialized payload $P$
\STATE $T\gets$ per-block $\max|m|$ and log-range of positive $v$ \COMMENT{blocks of 256}
\STATE $(c_m,c_v)\gets$ nearest 8-bit codes \COMMENT{joint: best of 4 pairs}
\STATE $(\widetilde m,\widetilde v)\gets\textsc{Decode}(T,c_m,c_v)$ \COMMENT{exact reader values}
\STATE $\mathcal A\gets\textsc{Certify}(\text{ideal-real bound},\,[n])$
\STATE $\mathcal A\gets\textsc{Certify}(\text{native bound }E,\,\mathcal A)$ \COMMENT{Eq.~\eqref{eq:accepted}}
\STATE $\mathcal X\gets([n]\setminus\mathcal A)\cup\{i:\text{standalone decode differs at }i\}$
\STATE $P\gets\text{hdr}\,\|\,T\,\|\,c_m\,\|\,c_v\,\|\,\mathbb{1}_{\mathcal X}\,\|\,(m_i,v_i)_{i\in\mathcal X}\,\|\,\text{SHA-256}$
\STATE \textbf{assert} a standalone parse of $P$ reproduces the decoded state
\STATE \textbf{return} $P$
\FUNCTION{$\textsc{Certify}(B,\mathcal U)$}
\STATE $\mathcal A\gets\emptyset$
\FOR{$K=64,256,1024$}
\STATE $\mathcal A\gets\mathcal A\cup\{i\in\mathcal U\setminus\mathcal A:\ B_i\text{ over }K\text{ intervals}\le\tau\}$
\ENDFOR
\STATE \textbf{return} $\mathcal A$
\ENDFUNCTION
\end{algorithmic}
\end{algorithm}

\subsection{Bounding the difference with a shared gradient}
\label{sec:shared}
Let $\alpha$, $b$, $\delta$, $c$, $e$ and $s<0$ be the realized FP32 coefficients of the native path for first-moment interpolation, variance decay, squared-gradient weighting, bias-root division, epsilon and the scaled step; ideally these are $1-\beta_1$, $\beta_2$, $1-\beta_2$, $\sqrt{1-\beta_2^{t+1}}$, $\epsilon$ and $-\eta/(1-\beta_1^{t+1})$. The exact-real center of the native path is
\begin{equation}
 \overline\Phi(w,m,v;g)
 =w+s\frac{(1-\alpha)m+\alpha g}
 {\sqrt{bv+\delta g^2}/c+e}.
 \label{eq:center}
\end{equation}
This expression uses the coefficients the implementation actually realizes rather than assuming that decimal metadata are represented exactly. The implementation also covers the pinned direct-division and rounded-reciprocal variants.

For an interval $I$ of signed gradients, let
$l=\min_{g\in I}|g|$ and $u=\max_{g\in I}|g|$. Define the lower denominators
\begin{align}
 a&=\sqrt{bv+\delta l^2},&
 \widetilde a&=\sqrt{b\widetilde v+\delta l^2},\\
 d&=a/c+e,&
 \widetilde d&=\widetilde a/c+e,
\end{align}
and the bounds
\begin{align}
 \Delta_n&=(1-\alpha)|m-\widetilde m|,&
 \Delta_d&=\frac{b|v-\widetilde v|}{c(a+\widetilde a)},\notag\\
 N&=(1-\alpha)|m|+\alpha u,&
 \widetilde N&=(1-\alpha)|\widetilde m|+\alpha u.
\end{align}
When both roots vanish, the variance difference also vanishes and $\Delta_d=0$. The two ways of splitting the quotient difference give
\begin{equation}
 B_I=\min\left\{
 \frac{\Delta_n}{d}+\frac{\widetilde N\Delta_d}{d\widetilde d},
 \frac{\Delta_n}{\widetilde d}+\frac{N\Delta_d}{d\widetilde d}
 \right\}.
 \label{eq:common}
\end{equation}
The shared gradient term $\alpha g$ cancels from the numerator difference, and the shared parameter $w$ cancels from the difference of centers, so $|\overline\Phi-\overline{\widetilde\Phi}|\le |s|B_I$. The bound scales with $|m-\widetilde m|$ and $|v-\widetilde v|$ rather than with the variation of the update over $I$. Ranging the two complete outputs separately loses both cancellations.

\subsection{Native rounding balls}
\label{sec:balls}
A \emph{ball} pairs an interval for an exact-real center with an absolute error radius. With $u_{32}=2^{-24}$ and $n_{\min}=2^{-126}$, a local operation whose unrounded argument magnitude is bounded by $M$ receives the radius
\begin{equation}
 R(M)=8u_{32}M+4n_{\min}.
 \label{eq:radius}
\end{equation}
The operation premise permits an outward double-to-FP32 conversion plus two adjacent FP32 values, including subnormal flushing; within the declared finite range, Equation~\eqref{eq:radius} conservatively covers that bracket. It is an assumption-backed arithmetic allowance, not a bound fitted to sampled errors.

Radii propagate through every operation. Under addition, input radii add; under multiplication, they contribute $|a|r_b+|b|r_a+r_ar_b$. Each operation then adds $R$ evaluated at a magnitude bound that includes those input errors. Square roots use the smaller of $\sqrt r$ and $r/(\sqrt x+\sqrt{\max(0,x-r)})$ when the latter is defined. Bias division, reciprocal multiplication, epsilon addition, quotient formation, step scaling and final parameter rounding all propagate radii, and the reciprocal path is recentered on the direct-division center with its coefficient discrepancy included.

The quotient requires a positive native denominator. In this profile it is at least $e/2$: the variance is nonnegative, the smallest supported bias root exceeds $0.03$, and the permitted negative square-root endpoint at zero is only a subnormal bracket whose propagation through bias scaling is negligible relative to $e\approx10^{-8}$; outward rounding around $e$ is also far below $e/2$. This independent lower-bound argument, not truncation of an arbitrary uncertain denominator, justifies using the larger of $e/2$ and the ball's lower endpoint.

With $r_I$ and $\widetilde r_I$ bounding the two native paths, a coordinate is accepted when
\begin{equation}
 E=\max_I\frac{|s|B_I+r_I+\widetilde r_I}{\eta}\le\tau.
 \label{eq:accepted}
\end{equation}
Every bound operation uses explicit outward-rounded double-precision intrinsics. Identical moment pairs take a deterministic identity shortcut, with the finiteness of the actual update checked separately. Equations~\eqref{eq:radius}--\eqref{eq:accepted} give a conditional enclosure under the pinned operation model; they are not a machine-checked verification of Torch or of arbitrary generated kernels.

\subsection{Fixed mixed partitions}
\label{sec:partition}
Within an interval, $B_I$ evaluates the denominators at the smallest magnitude $l$ and the numerators at the largest magnitude $u$. It is therefore loosest for intervals that span many orders of magnitude of $|g|$, which are precisely the intervals near zero, where a small $v$ makes the update most sensitive. At its first level, a uniform partition of $[-C,C]$ places every $|g|<C/32$ in a single interval on each side of zero (Figure~\ref{fig:design}b). The mixed partition instead combines evenly spaced linear edges with base-2 logarithmic magnitudes down to $2^{-40}C$, mirrored around zero. All edges are fixed in advance and never depend on future gradients or observed errors. Coordinates that remain unresolved are rechecked with 64, 256 and then 1024 intervals before they escape.

\subsection{Candidate codes and exact escapes}
\label{sec:codes}
Each 256-coordinate block stores three FP64 parameters: the largest $|m|$ and the logarithmic range of the positive $v$ values. The first moment uses a symmetric linear 8-bit code with an exact zero; the second uses an 8-bit logarithmic code over the block's positive range, with a separate code for zero. \emph{Independent} selection rounds each moment to its nearest decodable value. \emph{Joint} selection evaluates the four neighboring (floor and ceiling) code pairs and keeps the pair with the smallest ideal bound. Both selectors share the decoder, the guards and the acceptance rule, and both are judged on the values actually decoded. An escaped coordinate stores its source FP32 pair exactly and therefore restores the identical native update. Every header, table, code, mask and escape byte belongs to the measured payload, and the standalone decoder checks the entire serialized object.

\subsection{Implementation}
\label{sec:impl}
Both guards run as CUDA kernels compiled at run time with the NVRTC library shipped in the pinned PyTorch environment. Every bound operation uses explicit round-down or round-up double-precision intrinsics for addition, multiplication, division and square root, and each experiment records the source, PTX and cubin hashes and the compiler flags. The native enclosure is evaluated only for coordinates that pass the ideal-real guard. Decoding is deliberately redundant: a NumPy reader and an eager PyTorch reader reconstruct the moments in FP64 and must agree bitwise after the cast to FP32, and any coordinate on which they disagree is escaped. The payload header records the codec version, the contract parameters and the hash of the decoded state, so a reader rejects any payload whose bytes or decoded moments do not match. Restoration installs the decoded FP32 moments into the unmodified Torch optimizer state.

\section{Evaluation}
\label{sec:eval}
We answer four questions. \textbf{Q1}: Where do the savings come from? \textbf{Q2}: How does the guarded codec compare with equally guarded BF16 and blockwise 8-bit moments? \textbf{Q3}: Do complete checkpoints restore within the contract and continue training normally? \textbf{Q4}: What does guarding cost, and is joint code selection worth it?

\subsection{Methodology}
\label{sec:method}
\paragraph{Models and checkpoints.} We train four OPT-125M models~\citep{opt} (125,239,296 parameters; seeds 701, 707, 708 and 709) and checkpoint each at 10\%, 50\% and 100\% of training ($t=3{,}125$, $15{,}625$ and $31{,}250$ completed updates). We also train four ResNet-18 models~\citep{resnet} (11,181,642 parameters; seeds 701--704) on CIFAR-10~\citep{cifar} at resolution 224 for 20 epochs, or 12,500 updates, with checkpoints at $t=1{,}250$, $6{,}250$ and $12{,}500$. For mechanism studies we fix one OPT-125M checkpoint ($t=3{,}125$) and select exactly 1,000,000 coordinates in complete 256-coordinate blocks spread across the model, with one final partial block. No future gradient influences the selection, and the checkpoint, indices, code and compiled images are hashed.

\paragraph{Methods.} We compare five representations: source FP32; guarded BF16; guarded blockwise 8-bit, using official bitsandbytes~0.48.1 quantization with signed first-moment and unsigned second-moment dynamic maps and block size 256; and our 8-bit log codes with independent or joint selection (block size 256). All compressed methods use the same ideal and native guards, exact FP32 escapes and standalone serialized-byte decoders, and native bias coefficients are regenerated from each checkpoint's actual update counter. We compare checkpoint \emph{representations} consumed by the same Torch FP32 Adam; we do not train with the bitsandbytes optimizer recurrence.

\paragraph{Protocol.} Every condition restores both complete moment vectors. A common saved gradient tests the first restored update against the contract. Thereafter, each branch computes its own gradients for 100 updates from the same initial model, RNG state and data order, with minibatch indices matching exactly within each seed and checkpoint, and then evaluates the same 10,000 development examples. Feeding reference gradients to every branch would test repeated state decoding rather than training after restoration. Appendix~\ref{app:units} diagrams the protocol.

\paragraph{Units of analysis.} The independent units are the four trained seeds per architecture. Checkpoint fractions are nested within seeds, the five methods are paired within each seed and checkpoint, and the 10,000 development examples are repeated observations of one restored model. We therefore report descriptive means and seed ranges rather than confidence intervals, and never pool the two architectures, whose models, data and schedules differ. Development examples are paired by the frozen dataset row order because the evaluation files store no explicit example IDs. Timing uses separate cohorts in which the unit is a process launch (\S\ref{sec:cost}). All conditions are registered in advance and retained in the analysis, and all native experiments run on NVIDIA H100 GPUs.

\subsection{Where do the savings come from? (Q1)}
\label{sec:mechanism}
The ablation in Figure~\ref{fig:teaser}b fixes independent code selection, the gradient domain, the tolerance and the refinement budget, and varies only the bound and the partition. With separate ranges, 988,820 of the $10^6$ coordinates escape under either partition, and the payload is 126.63\% of raw FP32. Preserving the shared gradient lowers escapes to 439,362 (71.68\%) with the uniform partition, and adding the mixed partition lowers them to 5,324 (28.28\%). Every configuration spends the same 2,219,592 bytes on codes and metadata, and each escape adds 8 bytes, so the bound alone determines the payload. The two components are complementary: the mixed partition helps only once the shared gradient is preserved. The tighter bound is also cheaper. Warm encoding of the subset takes a median 0.161\,s with the shared-gradient bound and 0.435\,s with separate ranges (Appendix~\ref{app:extra}).

\paragraph{Implementation audit.} A source-matched H100 audit covers 1,048,576 scalar cases and 131,072 paired partitions and finds zero enclosure failures; a control with a deliberately missing radius triggers 899,311 detections. The compiled module contains the required directed arithmetic, and the source, PTX and cubin hashes and NVRTC flags are retained. This is finite implementation evidence under a stated arithmetic premise, not an all-input proof.

\subsection{Comparison with guarded baselines (Q2)}
\label{sec:baselines}
\begin{table}[t]
\caption{Complete moment payloads for the same one million trained OPT-125M coordinates ($\tau=0.01$, block size 256), relative to 8,000,000 raw FP32 bytes. Guarded rows share the same ideal-real and native guards. Plain blockwise 8-bit is smallest but violates the contract.}
\label{tab:subset}
\centering\small
\setlength{\tabcolsep}{4pt}
\begin{tabular}{@{}lrrr@{}}\toprule
Representation & Bytes & \% of FP32 & Escapes\\\midrule
Raw FP32 & 8,000,000 & 100.000 & ---\\
Serialized FP32 & 8,000,595 & 100.007 & 0\\
Plain BF16 & 4,125,281 & 51.566 & 0\\
Guarded BF16 & 4,125,333 & 51.567 & 0\\
Plain blockwise 8-bit & 2,158,768 & 26.985 & 0\\
Guarded blockwise 8-bit & 3,177,969 & 39.725 & 127,396\\\midrule
Log, independent (ours) & 2,262,239 & 28.278 & 5,324\\
Log, joint & 2,262,236 & 28.278 & 5,324\\
\bottomrule\end{tabular}
\end{table}

Table~\ref{tab:subset} compares complete payloads on the same subset. Under identical guards, log codes need 28.28\% of raw FP32 bytes: 45.16\% fewer than guarded BF16 and 28.82\% fewer than guarded blockwise 8-bit. Plain blockwise 8-bit storage is smaller still, but it violates the contract. On a fixed synthetic mixture of zero, logarithmic and uniform gradients in the declared domain, it exceeds $\tau$ on 2,469 of the million coordinates, with a maximum normalized error of 0.1294. Applying the same guards adds 127,396 escapes, after which no coordinate exceeds the tolerance on that witness (maximum 0.0098) and 896 high-precision point witnesses also pass. These finite witnesses supplement acceptance; they do not by themselves establish a continuous guarantee. Plain and guarded BF16 decode to the same values and show no exceedance on the witness; BF16 needs no escapes on this subset, so the two formats differ only in metadata. The official bitsandbytes CUDA decoder agrees exactly with our standalone payload parser. Independent and joint selection differ by three bytes.

\subsection{Complete checkpoints (Q2, Q3)}
\label{sec:full}
\begin{figure*}[t]
\centering
\includegraphics[width=\textwidth]{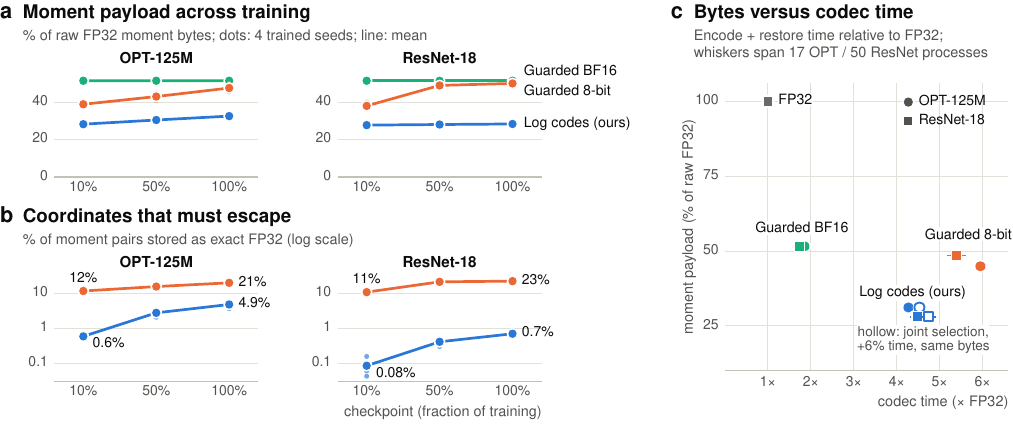}
\caption{Complete-checkpoint results. \textbf{(a)}~Moment payload at each checkpoint fraction; small dots are the four trained seeds and large dots their mean. FP32 is 100\%. Log codes use independent selection; joint selection differs by less than 0.002\% and would overlap. \textbf{(b)}~Moment pairs escaped to exact FP32. Guarded BF16 escapes at most 0.014\% of OPT-125M coordinates and none of ResNet-18's. \textbf{(c)}~Payload versus combined encode-and-restore codec time relative to FP32, from the separate timing cohorts (seed 701, 50\% checkpoint). Markers are geometric means over processes, whiskers span the process range, and hollow markers show joint selection.}
\label{fig:results}
\end{figure*}

\begin{table*}[t]
\centering\footnotesize
\caption{Complete-checkpoint sizes in MiB, averaged over four trained seeds at each training fraction. The moment payload includes serialization metadata and checksums; the complete container additionally holds the lossless model and the remaining checkpoint state. Payload and container are different denominators.}
\label{tab:full}
\setlength{\tabcolsep}{3.2pt}
\begin{tabular}{@{}lrrrrrr@{\hspace{12pt}}rrrrrr@{}}
\toprule
& \multicolumn{6}{c}{OPT-125M} & \multicolumn{6}{c}{ResNet-18}\\
\cmidrule(r{8pt}){2-7}\cmidrule{8-13}
& \multicolumn{3}{c}{Moment payload} & \multicolumn{3}{c}{Complete container} & \multicolumn{3}{c}{Moment payload} & \multicolumn{3}{c}{Complete container}\\
\cmidrule(r){2-4}\cmidrule(r{8pt}){5-7}\cmidrule(r){8-10}\cmidrule{11-13}
Method & 10\% & 50\% & 100\% & 10\% & 50\% & 100\% & 10\% & 50\% & 100\% & 10\% & 50\% & 100\%\\
\midrule
FP32 & 955.50 & 955.50 & 955.50 & 1993.32 & 1998.13 & 1997.27 & 85.31 & 85.31 & 85.31 & 165.78 & 166.02 & 166.44\\
Guarded BF16 & 492.81 & 492.81 & 492.80 & 1294.60 & 1299.44 & 1298.57 & 43.99 & 43.99 & 43.99 & 103.15 & 103.39 & 103.80\\
Guarded 8-bit & 372.49 & 412.03 & 455.61 & 1218.29 & 1259.65 & 1309.16 & 32.51 & 41.87 & 42.77 & 95.42 & 108.61 & 109.82\\
Log, indep.\ (ours) & 270.65 & 291.96 & 312.01 & 1021.20 & 1029.89 & 1056.70 & 23.73 & 24.01 & 24.26 & 78.92 & 79.33 & 80.17\\
Log, joint & 270.65 & 291.96 & 312.01 & 1021.20 & 1029.89 & 1056.69 & 23.73 & 24.01 & 24.26 & 78.92 & 79.33 & 80.17\\
\bottomrule
\end{tabular}
\end{table*}

Table~\ref{tab:full} and Figure~\ref{fig:results}a report the complete checkpoints. Averaged within seeds, log codes shrink the moment payload relative to FP32 by 71.67\%, 69.44\% and 67.35\% at the three OPT-125M fractions and by 72.18\%, 71.85\% and 71.56\% for ResNet-18. Relative to guarded BF16, the reductions are 45.08\%, 40.76\% and 36.69\% for OPT and 46.05\%, 45.41\% and 44.85\% for ResNet; relative to guarded blockwise 8-bit, they are 27.33\%, 29.08\% and 31.49\%, and 26.90\%, 42.64\% and 43.28\%.

Escapes explain these trends (Figure~\ref{fig:results}b). To meet the same contract, guarded blockwise 8-bit must escape 12\% of OPT coordinates at the first checkpoint and 21\% at the last, and 11\% to 23\% of ResNet coordinates. Log codes escape 0.6\% to 4.9\% and 0.08\% to 0.7\%, respectively. BF16 needs almost no escapes, but its 16-bit values alone already cost half of the FP32 bytes.

The moment payload is only part of a checkpoint. The complete serialized container also holds the lossless model and other state. With log codes it is 48.77\%, 48.46\% and 47.09\% smaller than the FP32 container for OPT and 52.39\%, 52.22\% and 51.83\% smaller for ResNet. A large reduction in the moment component therefore yields a smaller, though still substantial, reduction in the complete checkpoint, and we report both denominators.

\subsection{Restore fidelity and continuation (Q3)}
\label{sec:fidelity}
\begin{figure}[t]
\centering
\includegraphics[width=\columnwidth]{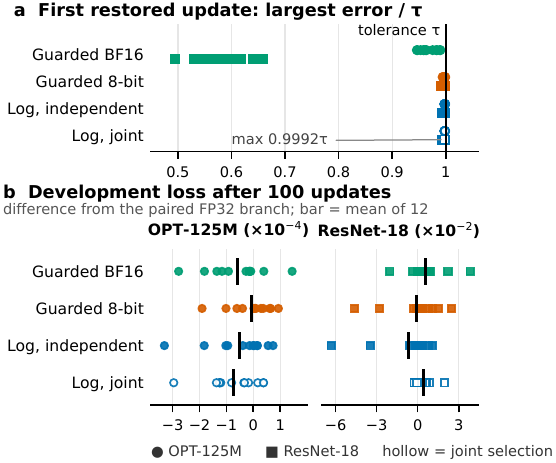}
\caption{Restore fidelity and continuation for the 96 guarded restores. \textbf{(a)}~Largest realized first-update error over all coordinates, per condition; every guarded condition stays below $\tau$, and the log-code and blockwise guards accept errors up to nearly the whole budget. \textbf{(b)}~Development-loss difference from the paired FP32 branch after 100 own-gradient updates, one marker per seed and checkpoint (note the different scales). Hollow markers show joint selection. Descriptive only; Table~\ref{tab:loss} in Appendix~\ref{app:extra} lists means and ranges.}
\label{fig:fidelity}
\end{figure}

All 120 conditions restore both complete moment vectors and pass native first-update acceptance and within-seed minibatch pairing. The largest realized first-update difference is 0.009992 learning rates for OPT and 0.009991 for ResNet, below $\tau=0.01$ in every condition (Figure~\ref{fig:fidelity}a), and the OPT audits record no point-enclosure violation for either the original or the restored update. The blockwise baseline's standalone decoder disagreed on one coordinate at OPT seed 701 (50\%) and one at seed 709 (100\%); each was stored as an exact escape, repacked and decoded successfully, and its bytes are included.

After 100 own-gradient updates, development losses remain close to the paired FP32 branch (Figure~\ref{fig:fidelity}b). For OPT, the mean difference between joint coding and FP32 is $-1.41\times10^{-4}$, $-4.18\times10^{-5}$ and $-3.68\times10^{-5}$ at the three fractions, and individual seeds take both signs at 50\% and 100\%. ResNet differences are larger for every compressed method, including BF16; for joint coding they average $+0.0087$, $+0.0012$ and $+0.0030$. These observations are descriptive: they concern development data, four seeds per architecture and 100 updates, and they establish neither held-out equivalence nor a bound on later trajectories.

\subsection{Cost, and whether joint selection pays (Q4)}
\label{sec:cost}
\begin{table*}[t]
\centering\small
\caption{Paired GPU codec timing at seed 701, 50\% checkpoint: arithmetic means of process means over 17 OPT-125M and 50 ResNet-18 processes, each with 20 warm-ups and 100 measured rounds per method. Peaks include the resident model and moments.}
\label{tab:timing}
\setlength{\tabcolsep}{4.2pt}
\begin{tabular}{@{}lrrrrr@{\hspace{14pt}}rrrrr@{}}
\toprule
& \multicolumn{5}{c}{OPT-125M ($N=17$ processes)} & \multicolumn{5}{c}{ResNet-18 ($N=50$ processes)}\\
\cmidrule(r{8pt}){2-6}\cmidrule{7-11}
& \multicolumn{3}{c}{Time (s)} & \multicolumn{2}{c}{Peak (GiB)} & \multicolumn{3}{c}{Time (s)} & \multicolumn{2}{c}{Peak (GiB)}\\
\cmidrule(r){2-4}\cmidrule(r{8pt}){5-6}\cmidrule(r){7-9}\cmidrule{10-11}
Method & Encode & Restore & Total & Encode & Restore & Encode & Restore & Total & Encode & Restore\\
\midrule
FP32 & 2.130 & 1.825 & 3.955 & 3.622 & 3.737 & 0.184 & 0.156 & 0.340 & 0.324 & 0.334\\
Guarded BF16 & 6.205 & 1.112 & 7.317 & 19.167 & 3.737 & 0.512 & 0.078 & 0.590 & 1.719 & 0.334\\
Guarded blockwise 8-bit & 20.864 & 2.664 & 23.528 & 17.350 & 3.737 & 1.588 & 0.245 & 1.833 & 1.516 & 0.334\\
Log, independent (ours) & 15.307 & 1.598 & 16.905 & 24.888 & 10.866 & 1.402 & 0.120 & 1.521 & 2.261 & 0.973\\
Log, joint & 16.315 & 1.597 & 17.912 & 27.806 & 10.866 & 1.491 & 0.119 & 1.610 & 2.522 & 0.973\\
\bottomrule
\end{tabular}
\end{table*}

\paragraph{Timing cohorts.} Separate cohorts of 17 OPT-125M and 50 ResNet-18 process launches time the codecs at seed 701 and the 50\% checkpoint. Every process runs all five methods in a frozen randomized order (Table~\ref{tab:timing}); the ResNet cohort alone comprises 30,000 attempts. Every encoding reproduces the accepted payload hash, and every restoration installs both complete moment vectors. Codec time is synchronized and covers internal transfers, both guards, payload packing and intrinsic checksums, standalone decoding, verification and GPU installation; initialization, the additional whole-payload SHA validation and container I/O are excluded and recorded separately. Ratios first compare method means within each process and are then summarized by their geometric mean and process range; raw rounds are never pooled as independent replicates.

\paragraph{Guarding is paid for at encode time.} Encoding with log codes takes 7.2$\times$ (OPT) and 7.6$\times$ (ResNet) as long as FP32 serialization, whereas restoration takes only 0.88$\times$ and 0.77$\times$ as long as FP32 restoration (Table~\ref{tab:timing}). Overall, independent log codes need 2.31$\times$ and 2.58$\times$ the combined codec time of guarded BF16 but only 0.72$\times$ and 0.83$\times$ that of guarded blockwise 8-bit, which they therefore beat on both bytes and time (Figure~\ref{fig:results}c). Encoding also raises peak allocation, to 24.9\,GiB for OPT versus 19.2\,GiB for guarded BF16 and 17.4\,GiB for guarded 8-bit, including the resident model and moments. These are component measurements on possibly shared hosts, not end-to-end checkpoint pauses or training throughput. Single executions within the 120 restore conditions show the same ordering (Appendix~\ref{app:extra}).

\paragraph{Joint selection does not pay for itself.} Beyond independent selection, joint selection saves 0.000526\%, 0.001022\% and 0.001365\% of OPT moment bytes at the three fractions, or 0.000971\% when fractions are averaged within seeds and seeds are weighted equally, and 11--195 bytes per ResNet checkpoint (0.000370\% on average; seed range 0.000285--0.000496\%). It is slower in every measured process: 1.060$\times$ the combined codec time for OPT (range 1.052--1.067) and 1.059$\times$ for ResNet (range 1.052--1.064), with 1.117$\times$ and 1.115$\times$ the encoding peak. Its ResNet restore-time ratio spans 0.977--1.007, so restoration offers no consistent advantage that would offset the extra encoding. Development losses do not favor it either. In ResNet, joint minus independent loss is positive at every seed (mean $+0.0113$, range $+0.0007$ to $+0.0216$; mean accuracy difference $-0.0026$), while the OPT seed contrasts take both signs (mean $-2.31\times10^{-5}$, range $-1.43\times10^{-4}$ to $+8.17\times10^{-5}$). Once the acceptance test is tight, independent rounding already captures essentially all of the available savings, so we recommend it.

\subsection{Bit width and tolerance (supporting study)}
\label{sec:sensitivity}
\begin{figure}[t]
\centering
\includegraphics[width=\columnwidth]{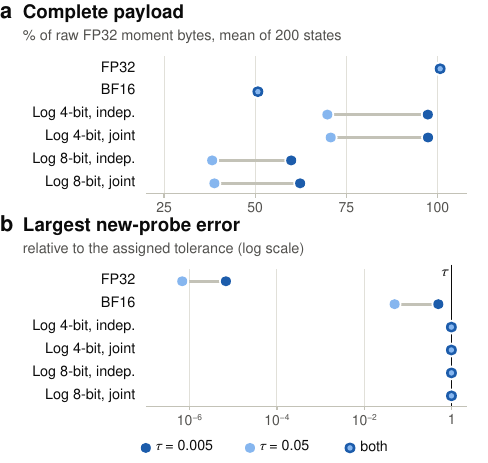}
\caption{Supporting study on 200 small logistic-optimization states with a binary64 guard. \textbf{(a)}~Mean complete payload relative to raw FP32 moments, including metadata and escapes. \textbf{(b)}~Largest new-probe error relative to the assigned tolerance; all formats also satisfy the 514 inherited probes. Zero tested violations are a numerical observation, not continuous native-operation certification.}
\label{fig:small}
\end{figure}
Our main experiments fix 8-bit codes and $\tau=0.01$. A supporting study varies both on 200 small logistic-optimization states (50 seeds, dimensions 128 and 512, sparsities 0 and 0.5), with independent and joint selection, 4- and 8-bit codes, and $\tau\in\{0.005,0.05\}$: 1,600 conditions in total (Figure~\ref{fig:small}; Appendix~\ref{app:small}). It uses an earlier binary64 format and a binary64 real-arithmetic guard rather than the native enclosure, so it controls for width and tolerance, not for the native guarantee. At $\tau=0.05$, independent and joint 8-bit payloads average 38.19\% and 38.81\% of raw FP32, against 50.73\% for BF16, and all 200 states compress and meet every probe. At $\tau=0.005$, BF16 is the smaller successful alternative to 8-bit log codes (50.73\% versus 59.87\%), and 4-bit codes save less than 3\%. Every format keeps its largest new-probe error below $\tau$ (at most $0.9888\tau$). Finite screens alone would not: on the inherited probes, 17-, 257- and 515-point screens leave 942, 371 and 15 failing conditions, and the 515-point screen leaves 16 failures on the disjoint probes, all of which the interval guard repairs.

\section{Related Work}
\label{sec:related}
\paragraph{Optimizer-state reduction.} Blockwise 8-bit optimizers~\citep{eightbit} and 4-bit optimizer states~\citep{fourbit} quantize Adam's moments during training, Adafactor factorizes second-moment storage~\citep{adafactor}, and GaLore projects gradients into a low-rank subspace~\citep{galore}. These methods change the optimizer's training-time representation and validate it empirically. ZeRO partitions optimizer state across workers~\citep{zero}, mixed-precision training assigns different numerical roles to storage and arithmetic~\citep{mixed}, and LoRA and QLoRA shrink the trainable state itself~\citep{lora,qlora}. We instead keep the optimizer recurrence unchanged and guard a checkpoint representation, coordinate by coordinate, against a declared update contract. The guard is representation-agnostic: we apply it to BF16 and to official bitsandbytes blockwise codes as well as to our own. This body of work also motivates explicit tensor scope and complete byte accounting, since a moment-payload reduction is not a reduction in all checkpoint or training memory.

\paragraph{Checkpointing systems.} CheckFreq~\citep{checkfreq} and Gemini~\citep{gemini} make frequent checkpointing and fast recovery practical, and Check-N-Run~\citep{checknrun} reduces checkpoint size and bandwidth for recommendation models. LC-Checkpoint~\citep{lccheckpoint} and ExCP~\citep{excp} compress checkpoints lossily, the latter by jointly shrinking weights and momentum. Lossy approaches judge compressed state by the accuracy recovered after restart rather than by bounding the restored optimizer update. A guarded codec is complementary and could serve as the moment component of such systems.

\paragraph{Floating-point and interval analysis.} Floating-point analysis separates real inequalities, rounded evaluation and implemented operation contracts~\citep{goldberg,higham2002}. Interval arithmetic and its dependency problem are classical~\citep{moore2009}, and FPTaylor combines symbolic error analysis with rigorous optimization and can emit proof certificates~\citep{fptaylor2015}. We do not introduce a general round-off analysis or a machine-checked proof system; our contribution is a problem-specific enclosure that retains the shared gradient in an Adam update difference and couples it to serialized acceptance. Stochastic rounding offers a separate, probabilistic approach~\citep{rounding}; our acceptance and decoding are deterministic.

\section{Discussion and Scope}
\label{sec:limits}
\paragraph{Choosing a representation.} Figure~\ref{fig:results}c suggests a simple rule. When encode time dominates, for example with very frequent checkpoints, guarded BF16 saves 48\% of moment bytes at 1.7--1.8$\times$ the FP32 codec time. When storage or transfer dominates, guarded log codes save 69--72\% on average at 4.3--4.5$\times$, and they beat guarded blockwise 8-bit on both bytes and time. Restoration is not the bottleneck for log codes, which restore faster than FP32 on both models.

\paragraph{Lessons for guarded codecs.} Three lessons may carry over to other codecs for training state. First, certify the consumer's computation rather than the stored values: training sees the restored update, not the moment error, and the same guard applies unchanged to BF16 and blockwise codes. Second, when certifying the difference between two computations that share inputs, keep the inputs shared. On our checkpoint, the shared gradient together with a partition that resolves small gradients moved the payload from 126.6\% to 28.3\% of raw FP32, whereas the choice between independent and joint codes changed it by less than 0.002\%. Third, account for every byte and every exception. Exact escapes make the contract total, and charging them to the payload is what exposes both the failure of separate ranges and the escapes that make guarded blockwise 8-bit larger than log codes.

\paragraph{What the guarantee covers.} The enclosure holds under the pinned, restricted operation model of \S\ref{sec:problem}. Finite audits, successful probes and inspected instructions corroborate that model on the implementation; extending the guarantee to a formally verified framework binary and arbitrary generated kernels is a natural next step.

\paragraph{After the first update.} The shared-gradient cancellation holds only at restoration, when both moment pairs receive the same parameter, gradient and metadata. Afterwards each restored model generates its own gradients, so the cancellation no longer yields a trajectory bound, and forcing every branch to reuse reference gradients would test a different computation. The 100-update continuation therefore observes the consequence of the actual intervention; passing the first-update audit and observing a later loss difference are compatible outcomes, not competing criteria for the same claim.

\paragraph{Bytes and memory.} Payload and container bytes answer different systems questions: at the ResNet 50\% checkpoint, the moment payload falls from 85.31 to 24.01\,MiB but the complete container only from 166.02 to 79.33\,MiB. The restored moments are consumed by the unchanged FP32 Torch update path, so the codec drops into existing training loops; a permanently compressed optimizer recurrence is a natural extension. Peak allocation while constructing, validating and installing the state is a separate quantity, higher for log codes than for the baselines.

\paragraph{Systems scope.} Our timings isolate the codec boundary. Intrinsic payload checksums are inside the timed interval, while the additional identity SHA check, full-container serialization, filesystem transfer and resumed training are outside it; end-to-end checkpoint pipelines are future work. Processes may share hosts, allocation waves and competing work, and we record that resource context with every measurement.

\paragraph{Statistical scope and scale.} Each architecture contributes four trained seeds, and we report each architecture separately rather than pooling them into a common utility effect. Development observations are paired by row order and reported as descriptive differences. The results span OPT-125M and ResNet-18; billion-parameter and sharded checkpoints are the natural next scale. The supporting small-state study in Appendix~\ref{app:small} uses a different source precision and numerical guard.

\section{Conclusion}
Certifying compressed optimizer state is a dependency problem: the source and restored updates share their gradient and parameter, and a certificate that ignores this is too loose to compress anything. Bounding the difference directly, with explicit FP32 rounding allowances, fixed mixed partitions and exact escapes, turns a 126.6\% ``compressed'' payload into 28.3\% of raw FP32 under the same per-coordinate contract. Across 120 complete OPT-125M and ResNet-18 restores, guarded log codes are on average 40.8--45.4\% smaller than equally guarded BF16 and 29.3--37.6\% smaller than equally guarded blockwise 8-bit moments, every first restored update stays within tolerance, and the price is paid in encoding time and memory. Independent per-moment rounding suffices; joint code search does not pay for itself. Natural next steps are sharded, billion-parameter checkpoints and an optimizer recurrence that stays compressed between checkpoints.

\section*{Reproducibility Statement}
The artifact retains the pinned source files, frozen protocols, original inputs, counterexamples, every comparison condition and the source hashes. Predictions, nested timing repetitions, independent seeds and full-model experiments are accounted separately, and no measured subset or numerical audit is counted as completed model training or quality confirmation.

\section*{AI Usage Disclosure}
We used ChatGPT (OpenAI) to refine the manuscript's language and formatting, and Claude (Anthropic) to assist with figure preparation. AI assistance was limited to these presentation-related tasks and did not extend to research ideation, literature review, experimental design, research code generation, or the development of technical content. We reviewed and approved all AI-assisted edits and figures and take full responsibility for the final manuscript.

\clearpage
\bibliography{references}

\bibliographystyle{mlsys2025}

\clearpage
\onecolumn
\appendix
\section{Paired Restoration and Measurement Units}
\label{app:units}
\begin{figure}[!ht]
\centering
\includegraphics[width=\textwidth]{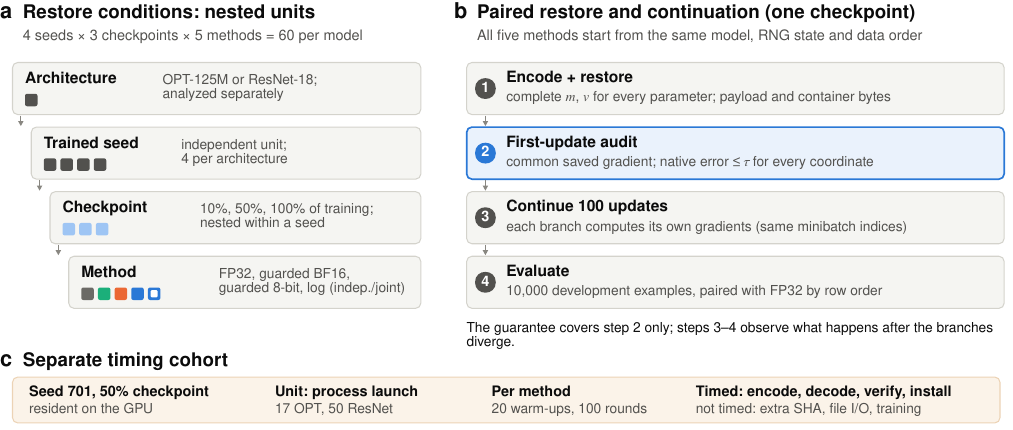}
\caption{Paired restoration and its measurement units. \textbf{(a)}~OPT-125M and ResNet-18 each contribute four independently trained seeds; three checkpoint fractions are nested within each seed, and all five representations are paired within each checkpoint. \textbf{(b)}~Every representation restores complete moment tensors, shares the first saved-gradient audit, and then continues with its own gradients from the same starting model, RNG state and data order. Ordered development evaluation is descriptive, not held-out quality equivalence. \textbf{(c)}~The separate timing cohorts contain 17 OPT and 50 ResNet processes. Their synchronized codec intervals exclude the additional whole-payload SHA check, disk transfer, full-container serialization and later training.}
\label{fig:units}
\end{figure}

Figure~\ref{fig:units} makes the nested experimental units, the paired own-gradient continuation and the separate timing boundary explicit. It summarizes the protocol of \S\ref{sec:method} and adds no measurement or performance claim. Repeated rounds, methods and checkpoint fractions do not increase the number of independent seeds or timing processes.

\section{Supporting Small-State Control Study}
\label{app:small}
This appendix details the supporting study summarized in \S\ref{sec:sensitivity} and Figure~\ref{fig:small}. It provides broader width and tolerance controls on small optimizer states. It uses an earlier 64-coordinate format whose source moments are binary64; its FP32-escape extension checks rounded FP32 escapes and retains an exact FP64 fallback when necessary. Its source precision, format and numerical guard therefore differ from the native study in the main text.

\paragraph{Design.} The experiment comprises 50 independent training seeds, dimensions 128 and 512, and sparsities 0 and 0.5, giving 200 logistic-optimization states saved after $t=64$ updates. Independent and joint codes, 4- and 8-bit widths and tolerances $\tau=0.005$ and $0.05$ give 1,600 conditions. Every state, tolerance and outcome is retained. Three serialization variants produce 4,800 payload records; these repetitions do not create additional independent seeds.

\paragraph{Reproduction and probes.} All 1,600 guarded payloads are byte-identical to the archived artifact. The executing decoder reconstructs archived transcendental values before evaluation, accounting for platform-level binary64 differences. The 514 inherited and 1,536 disjoint gradient probes use the same binary64 update consumer for all storage methods, and the new probe magnitudes were fixed independently of observed errors. NumPy supplies portable state and serialization operations~\citep{numpy2020}. These are storage interventions under a common numerical consumer, separate from the native study.

\paragraph{Guard.} The real-arithmetic inequality has the same quotient-difference structure as Equation~\eqref{eq:common}, with the bias-corrected coefficients of this format. It is evaluated in binary64 over 512 magnitude intervals with a specified margin, not with explicit directed rounding.

\begin{table}[!ht]
\caption{Complete serialized bytes divided by two raw FP32 moment arrays ($8d$), with the largest new-probe error relative to $\tau$. Each row retains all 200 states. Guarded rows use approximate FP32 escapes with an explicit exact-FP64 fallback. Both baseline formats pass every probe budget at both tolerances.}
\label{tab:fp32-storage}\centering\small
\begin{tabular}{@{}lrrrr@{}}\toprule
Method & Bits & $\tau$ & Bytes / FP32 & Max.\ error / $\tau$\\\midrule
Independent & 4 & 0.005 & 0.9735 & 0.9832 \\
Independent & 4 & 0.05 & 0.6979 & 0.9888 \\
Independent & 8 & 0.005 & 0.5987 & 0.9873 \\
Independent & 8 & 0.05 & 0.3819 & 0.9888 \\
Joint & 4 & 0.005 & 0.9740 & 0.9832 \\
Joint & 4 & 0.05 & 0.7071 & 0.9870 \\
Joint & 8 & 0.005 & 0.6232 & 0.9873 \\
Joint & 8 & 0.05 & 0.3881 & 0.9835 \\
\midrule
BF16 & 16 & 0.005 & 0.5073 & 0.4942\\
BF16 & 16 & 0.05 & 0.5073 & 0.0494\\
FP32 & 32 & 0.005 & 1.0073 & $6.8\times10^{-6}$\\
FP32 & 32 & 0.05 & 1.0073 & $6.8\times10^{-7}$\\
\bottomrule\end{tabular}
\end{table}

\paragraph{Results.} At $\tau=0.05$, mean independent and joint 8-bit payload ratios are 0.3819 and 0.3881, while complete BF16 uses 0.5073; all 200 states compress and meet all probes. At the tighter tolerance, BF16 is the smaller successful alternative. Descriptive 95\% intervals that resample the four states of each seed together over the 50 seeds are $[0.3807,0.3831]$ and $[0.3866,0.3896]$ at $\tau=0.05$, versus $[0.5905,0.6069]$ and $[0.6152,0.6315]$ at $\tau=0.005$. These pointwise intervals do not establish downstream training noninferiority.

\paragraph{Finite screens are not enough.} The interval guard repairs failures that finite screens miss. On the inherited probes, 17-, 257- and 515-point screens leave 942, 371 and 15 failing conditions, and the 515-point screen leaves 16 failures on the disjoint probes; the interval guard passes every retained condition. Selected witnesses recomputed at 80 decimal digits show no guard shortfall. The original exact-FP64-escape format's 50.23\% ratio uses binary64 arrays as its denominator and is not substituted for FP32 savings.

\paragraph{Reachable histories.} A separate study completes all 10,000 fixed seed keys over history lengths 1, 10 and 1,000 at dimension 16. Its 111 aggregate method, width, block and length conditions have zero confirmed next-update tolerance violations and zero recorded high-precision witness failures, and future gradients are generated only after payloads are frozen. This result validates the earlier guarded native path with its escapes; it is neither an independent replication of the new shared-gradient implementation nor a trained-model quality experiment.

\section{Additional Measurements}
\label{app:extra}
\paragraph{Warm encoding on the subset.} Table~\ref{tab:native-warm} separates warm encoding from cold compilation on the one-million-coordinate subset. Five randomized-order nested repetitions per variant run in one process after cold compilation and warm-up; these correlated observations are not process-level performance confirmation, and cold first-call timings are not used to rank the selectors. Multiplying subset throughput by model size does not measure complete-checkpoint cost.

\begin{table}[!ht]
\caption{Warm encode medians in seconds on the same one million OPT-125M coordinates.}
\label{tab:native-warm}
\centering\small
\begin{tabular}{@{}lrr@{}}\toprule
Bound & Independent codes & Joint codes\\\midrule
Separate ranges & 0.43545 & 0.44392\\
Shared gradient, mixed partition & 0.16056 & 0.16320\\
\bottomrule\end{tabular}
\end{table}

\paragraph{Single-execution costs in the restore conditions.} These whole-condition observations are descriptive and distinct from the registered timing cohorts of \S\ref{sec:cost}. For OPT-125M across the three fractions, joint coding's geometric-mean encode-time ratios are 0.857--0.889 relative to guarded blockwise 8-bit and 2.373--2.506 relative to guarded BF16, and its measured peak allocation is 1.552--1.644 and 1.435--1.467 times those baselines. Against independent log coding, joint selection takes 1.054--1.064 times the encoding time and 1.116--1.119 times the peak allocation. Mean joint encode times are 16.15, 17.20 and 17.78\,s, with mean peaks of 28.11, 27.88 and 27.63\,GiB. These peaks include the resident model and moments and the encode, decode, verification and continuation interval after an allocator reset; they exclude earlier model construction and are not persistent optimizer-memory reductions. For ResNet-18, joint encoding takes 1.011--1.034 times the time and 1.1158--1.1164 times the allocated peak of independent selection; its encode-time ratios are 1.584--1.718 relative to guarded BF16 and 0.564--0.590 relative to guarded blockwise 8-bit, with higher allocated peaks than both.

\begin{table}[!ht]
\caption{Development-loss difference from the paired FP32 branch after 100 own-gradient updates: mean and [min, max] over 4 seeds $\times$ 3 checkpoints. Descriptive only; positive values mean higher loss than FP32.}
\label{tab:loss}
\centering\small
\setlength{\tabcolsep}{3.5pt}
\begin{tabular}{@{}lrr@{}}\toprule
Method & OPT-125M ($\times10^{-4}$) & ResNet-18 ($\times10^{-2}$)\\\midrule
Guarded BF16 & $-0.58$ [$-2.78$, $1.45$] & $0.59$ [$-2.04$, $3.85$]\\
Guarded 8-bit & $-0.07$ [$-1.91$, $0.93$] & $-0.10$ [$-4.63$, $2.48$]\\
Log, independent & $-0.50$ [$-3.31$, $0.73$] & $-0.69$ [$-6.33$, $1.07$]\\
Log, joint & $-0.73$ [$-2.96$, $0.38$] & $0.43$ [$-0.26$, $1.96$]\\
\bottomrule\end{tabular}
\end{table}

\paragraph{Development details.} Table~\ref{tab:loss} summarizes the paired development-loss differences plotted in Figure~\ref{fig:fidelity}b. For ResNet-18, joint minus independent loss averages $+0.0349$, $-0.0015$ and $+0.0004$ at the three fractions, and joint minus FP32 accuracy averages $-0.00066$. For the timing cohorts, start and end resource context describes contention where available; the first OPT process predates the resource observer. The timing observations add no trained-seed units or held-out quality data.

\end{document}